# Multi-Phase Automated Segmentation of Dental Structures in CBCT Using a Lightweight Auto3DSeg and SegResNet Implementation

Dominic LaBella [1] [0000-0003-1713-9538], Keshav Jha [2], Jared Robbins[1] [0000-0003-1724-5242], Esther Yu[1]

[1] Department of Radiation Oncology, Duke University Medical Center, Durham, NC 27710, USA
[2] Duke University, Durham, NC 27710, USA

**Abstract.** Cone-beam computed tomography (CBCT) has become an invaluable imaging modality in dentistry, enabling 3D visualization of teeth and surrounding structures for diagnosis and treatment planning. Automated segmentation of dental structures in CBCT can efficiently assist in identifying pathology (e.g., pulpal or periapical lesions) and facilitate radiation therapy planning in head and neck cancer patients. We describe the DLaBella29 team's approach for the MICCAI 2025 ToothFairy3 Challenge, which involves a deep learning pipeline for multi-class tooth segmentation. We utilized the MONAI Auto3DSeg framework with a 3D SegResNet architecture, trained on a subset of the ToothFairy3 dataset (63 CBCT scans) with 5-fold cross-validation. Key preprocessing steps included image resampling to 0.6 mm isotropic resolution and intensity clipping. We applied an ensemble fusion using Multi-Label STAPLE on the 5-fold predictions to infer a Phase 1 segmentation and then conducted tight cropping around the easily segmented Phase 1 mandible to perform Phase 2 segmentation on the smaller nerve structures. Our method achieved an average Dice of 0.87 on the ToothFairy3 challenge out-of-sample validation set. This paper details the clinical context, data preparation, model development, results of our approach, and discusses the relevance of automated dental segmentation for improving patient care in radiation oncology.

## 1 Introduction

Cone-beam computed tomography (CBCT) has revolutionized dental imaging over the past two decades, overcoming the limitations of 2D panoramic radiography and providing accurate multiplanar visualization of maxillofacial structures (1–3). CBCT's ability to produce high-resolution 3D images enables improved detection of dental pathologies. Notably, the most common pathologic conditions involving teeth, inflammatory lesions of the pulp and periapical areas, can be visualized more reliably with CBCT than with conventional radiographs (3). Lesions confined to cancellous bone that might be missed on intraoral X-rays are often evident on CBCT, leading to greater diagnostic accuracy, as shown by Jaju et al, where CBCT detected periapical lesions with ~61% accuracy vs ~39-44% for digital or film radiographs (3). Such 3D information is clinically valuable for endodontic evaluation and treatment planning, allowing clinicians to assess the true extent of pulp chamber infections, periapical cysts, or granulomas and to plan surgical interventions accordingly. The broad adoption of dental CBCT reflects its utility in implantology, orthodontics, and oral surgery, as well as in baseline dental evaluations for oncology patients (2).

In patients with head and neck cancer, dental health management before and after radiation therapy (RT) is critical (4,5). Irradiation can compromise oral health by reducing salivary flow and blood supply to the jaws, leading to higher risk of dental caries, periodontal disease, and osteoradionecrosis (ORN) of the jaw (6). ORN, a severe complication where irradiated bone fails to heal, has an incidence of roughly 1-9% in RT patients and occurs much more frequently in the mandible (~85% of cases) than the maxilla (6). The risk of ORN is strongly dose-dependent, rising from <6% at doses below 40 Gy to ≥20% at doses above 60 Gy (6). Clinical practice guidelines therefore recommend proactive dental management: for example, teeth anticipated to receive very high radiation doses may be extracted prophylactically (common thresholds are ≥70 Gy in the maxilla or ≥60 Gy in the mandible for considering extraction) to prevent ORN (7). Even with such measures, post-RT dental extractions or infections can precipitate ORN, so accurately mapping radiation dose to each tooth is important in predicting risk. Per the classic Marx protocol, hyperbaric oxygen is used as an adjunct around dental surgery in irradiated jaws, typically ~20 pre-extraction and 10 post-extraction sessions at ~2.4 times atmospheric pressure for 90 min, and for established ORN as staged therapy beginning with ~30 sessions followed by limited debridement and ~10 additional dives to promote angiogenesis and wound healing (8).

In current practice, radiation oncologists and dental specialists collaborate to evaluate teeth in or near high-dose regions using CT imaging and clinical exam. However, this process is largely manual and qualitative. Automated tooth segmentation on planning CBCT scans could greatly enhance this workflow by providing precise tooth contours for dose-effect analysis (6). Prior work by Thariat et al. introduced an atlas-based auto-segmentation of dental structures ("Dentalmaps"), demonstrating that using automatically segmented teeth to estimate per-tooth radiation dose was significantly more accurate (within 2 Gy in 75% of cases) than visual estimation without contours (within 2 Gy in only 30% of cases) (6). Such tools improve communication between radiation oncologists and dentists and help identify teeth at highest risk for complications (6).

The MICCAI ToothFairy3 Challenge was organized to advance fully automated, multi-class segmentation of dental and maxillofacial structures in CBCT volumes (9–12). The challenge dataset provides 3D CBCT scans with detailed annotations of 77 anatomical labels, including all teeth (with individual tooth identifiers), dental restorations, pulp canals, nerves, and surrounding tissues (9–12).

In this paper, we present our team's approach and results in the ToothFairy3 Challenge 2025. We aimed for a solution that is robust and computationally efficient, leveraging the MONAI Auto3DSeg framework to automatically configure a deep neural network for the task (13). We describe our data preparation (focusing on a subset of the training data with full dental field-of-view), model training with 5-fold cross-validation, Multi Label STAPLE ensemble fusion method, post-processing techniques, and cropping of an initial "Phase 1" prediction to perform a focused "Phase 2" inference for final predictions on the smaller nerve structures. We also discuss the clinical relevance of the results and how such automated segmentations can be integrated into radiation therapy planning to help reduce dental complications.

## 2      Methods

### 2.1    Dataset Selection

*Phase 1 Dataset*
We utilized the ToothFairy3 challenge training dataset, which consists of 532 CBCT scans annotated with 77 substructure classes (9–12). Due to region of interest (ROI) and training time considerations, we restricted "Phase 1" training from the entire provided set to just the "Set B" subset, comprising 63 CBCT image-label pair cases as seen in **Figure 1A**. This subset has a broader scanning range (head CBCT images capturing all teeth) compared to "Set A" (n = 417) and "Set C" (n = 52) from the full dataset, that were cropped and frequently missing challenge evaluated substructures. Each "Set B" image had an isotropic voxel size of approximately 0.3 mm$^3$, a median voxel volume of [168, 362, 371], and was provided with a corresponding segmentation label map for a subset of the 77 substructures. The 77 substructure classes in the challenge provided dataset were consolidated into 46 substructures as described by the challenge organizers (12). This full-sized dataset was used for initial training of a Phase 1 multi-class automated segmentation model.

*Phase 2 Dataset*
We cropped the full-sized Phase 1 dataset image and reference standard label pairs to a ROI around the mandible (label = 1). The ROI was determined based on the reference standard mandible label by identifying the point (x, y, z) representing the most anterior voxel for the mandible, then expanding laterally (x) by -110 and +110 voxels (ensuring staying within boundaries of full-sized image); then expanding posteriorly (y) by +100 voxels; then expanding superiorly from the most inferior mandibular point by 90 voxels. The final result of this cropping is demonstrated in **Figure 1B.** Note that this reduced the total Phase 2 image sizes by about 60 times compared to the full-sized Phase 1 images, dramatically reducing the image data needed to be evaluated during Phase 2 model training.

All image and labels files were provided as de-identified and standardized Neuroimaging Informatics Technology Initiative (NIfTI) format.

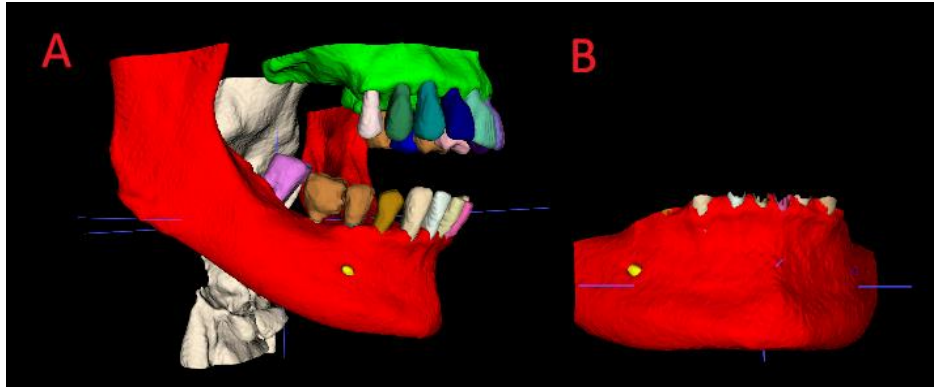

**Fig. 1.** Panel A demonstrates an example of an unaltered, ToothFairy3, "Set B", Phase 1 training set case 3D multi-label representation demonstrating the majority of challenge substructures. Note that the brown "Crown" structure in in place of multiple teeth for this case. Panel B demonstrates an example of the corresponding cropped Phase 2 training set case 3D multi-label representation, which isolates the region of interest surrounding the Right Incisive Nerve, Left Incisive Nerve, and Lingual Nerve.

## 2.2    Preprocessing and Label Conversion

For Phase 1 alone, we resampled all CBCT images to a uniform resolution of [0.6, 0.6, 0.6] mm$^3$ isotropic spacing to standardize the input size for the network and to account for hardware memory and processing speed limitations for the larger image sizes seen with native resolution of [0.3, 0.3, 0.3] mm$^3$. Intensity values (in Hounsfield Units) were clipped to [-1000, 3800] for both Phase 1 and Phase 2, which encompasses the range from air to the highest densities of enamel/metal artifacts in the scans. We also applied systematic label remapping to simplify the segmentation task. In the original challenge labels, certain structures had high integer label values or were subdivided into many small categories (for example, each tooth's pulp cavity was labeled separately with IDs 111-148). We merged these into a consolidated label set for model training. Specifically, all pulp and periapical lesion labels were collapsed into a single "pulp" class, and the incisive and lingual canals (original labels 103-105) were re-indexed to fit within the 0-46 range (while preserving distinct labels for evaluation, background = 0). The remapping reduced sparsely represented classes and ensured that the network's output channel count did not skip any integer label values, to align with Auto3DSeg requirements. After conversion, the training label maps included: background, 32 individual tooth labels (upper and lower teeth 1-16), the mandibular and maxillary jaw bone, dental restorations (implants, crowns, bridges; none included in challenge ranking metrics), bilateral inferior incisive nerves, alveolar canals, the lingual nerve, maxillary sinuses, pharynx, and a combined pulp category.

## 2.3    Model and Training

We adopted the Auto3DSeg automated segmentation pipeline implemented in MONAI, which streamlines model configuration and hyperparameter tuning (13,14). The chosen backbone model was SegResNet, a 3D residual UNet-like convolutional network known to perform well in medical image segmentation challenges (14,15). Auto3DSeg initialized a SegResNet with default encoder-decoder structure and optimized training settings based on our data. The encoder used five ResNet blocks with instance normalization, and the downsampling included five stages with 1, 2, 2, 4, and 4 convolutional blocks, respectively, similar to our prior experience (16).  We sectioned the training and inference into two phases.

*Phase 1*

We trained the Phase 1 model for 500 epochs on the 63 full-sized training images and associated reference standard multi-class labels, as seen in **Figure 1A,** using a random 5-fold cross-validation (each fold held out ~20% of cases for validation)**.** The training objective was a combined Dice + Cross-Entropy loss (implemented as DiceCELoss in MONAI, with equal weighting) computed over all foreground classes. Notably, we excluded background voxels from the Dice term to focus the loss on meaningful structures. We enabled automatic mixed precision (AMP) to accelerate training, and used a batch size of 1 (one 3D volume per GPU iteration) and a region of interest (ROI) size of [192, 192, 128] due to memory and time constraints. Data augmentation included random intensity scaling and shifting, and slight rotations, as configured by Auto3DSeg's defaults, to improve generalization. Random flipping data augmentation was specifically disabled to prevent inaccurate training of contralateral dental substructures. The learning rate of 0.0003 had a weight decay of 0.00005 with the use of the AdamW optimizer. Data caching and the use of multiple workers was not able to be performed due to repeated crashing during model training due to RAM, CPU, and I/O overload. All model training was conducted on a single NVIDIA RTX 4090 GPU (24 GB of VRAM available), but only utilized approximately 8 GB of VRAM during training. These decisions were made due to the longer training times and more frequent training crashes associated with using larger networks and ROIs associated with larger available VRAM utilization. Each fold's training took approximately 6-10 hours. Our use of a single-GPU workstation contrasts with some recent Auto3DSeg challenge solutions that leveraged multi-GPU servers (8 × A40 GPUs) that performed more folds using different networks including Swin UNETR and DiNTS (15,17,18). In our methodology, we focused on the single SegResNet model approach due to resource limitations; thereby accepting a longer training time, smaller ROI sizes, and smaller SegResNet network structures. This decision was made due to the superior performance of SegResNet compared to Swin UNETR and DiNTS as described by Myronenko et al (15).

*Phase 2*

Phase 2 included training on the cropped version of the Phase 1 images and associated reference standard labels that focus on a small region of interest surrounding the difficult to infer lingual nerve as seen in **Figure 1B**. These identical expansions were used during the Phase 2 inference stage. Phase 2 training was similar to Phase 1, except for an ROI of [221, 101, 91] voxels and no resampling to [0.6, 0.6, 0.6] mm$^3$. The VRAM used was limited to 3 GB. The purpose for retaining with the [0.3, 0.3, 0.3] mm$^3$ higher resolution images was to try and segment the small nerve structures which have a very small tubular diameter.

### 2.4    Inference and Ensemble Fusion

First, we ran inference with each of the 5-fold SegResNet Phase 1 models on each test CBCT, yielding five candidate label maps. We then applied a label fusion algorithm to combine these outputs into one consensus segmentation. In particular, we used the Multi-Label Simultaneous Truth and Performance Level Estimation (STAPLE) (19). STAPLE is an expectation-maximization algorithm that weighs each input segmentation by its estimated accuracy and computes a probabilistic "true" segmentation (19). We chose STAPLE because it is well-suited for fusing multiple label maps and can handle the multi-class nature of our problem (via an extension to Multi-Label STAPLE). The five model outputs were given to the STAPLE filter implemented in SimpleITK, which produced a fused label map.

After STAPLE ensemble was performed, we converted any aberrant class predictions touching the larynx (label = 7) to the larynx. We also removed any mandible instance lesions that had a volume less than 200,000 voxels, which would indicate false positive instance lesions. **Figures 1-3** illustrates our overall Phase 1 workflow, from preprocessing the input CBCT to generating the final ensembled segmentation. **Figure 4** represents the cropped ROI for Phase 2 inference as described in section 2.3. Phase 2 inference was conducted similarly to the steps seen in Phase 1 inference, except for the use of a set of five different cross-validation models that were trained on the cropped dataset as shown in **Figure 1B.** Note that **Figures 1-4** represent an in-sample case from Fold 1, and therefore give an artificially high inference performance appearance. Due to the small number of selected training set cases (n = 63), no "Set B" cases were left out-of-sample for internal testing for quantitative or qualitative analysis prior to challenge submission. This was to done to try and create the most generalizable model possible from the largest training set possible.

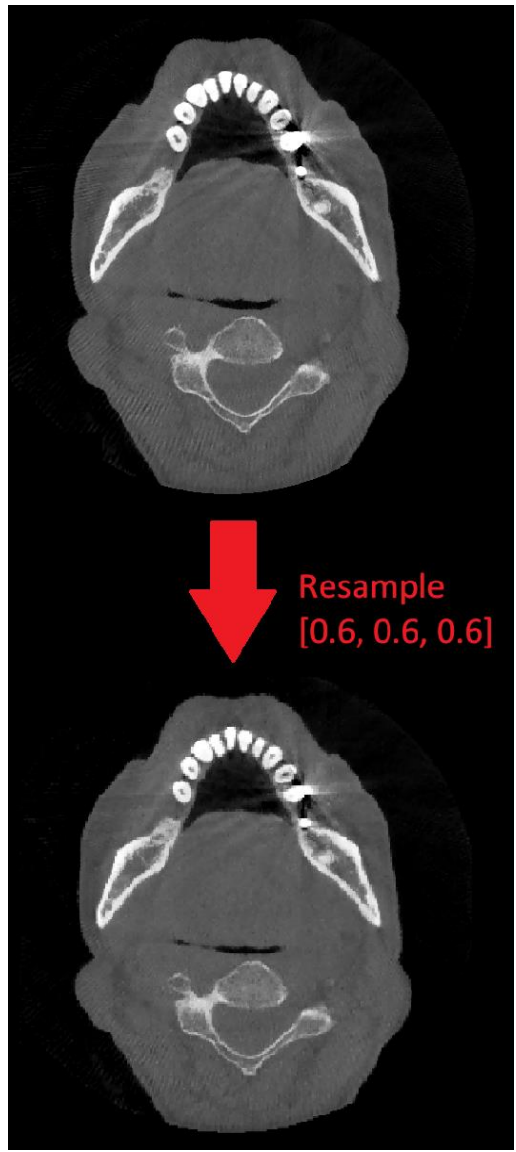

**Fig. 2.** Axial CBCT images demonstrating that during Phase 1 training and inference, we first preprocess the input CBCT using resampling of native resolution of about [0.3, 0.3, 0.3] mm$^3$ to consistent [0.6, 0.6, 0.6] mm$^3$ for all cases.

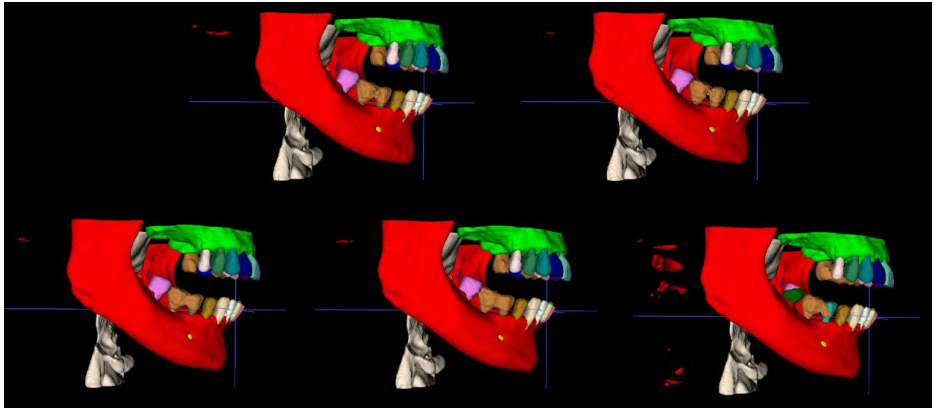

**Fig. 3.** For Phase 1 inference, we apply the trained SegResNet models to predict multi-class label maps (folds 1–5) and convert back to the native image resolution as seen in these 3D multi-label representations. Note that this is an in-sample validation case from training fold 1 (top left). Therefore folds 2-5 (top middle, top right, bottom left, bottom right) had this case in the training sets and had artificially high performance during this inference sanity check.

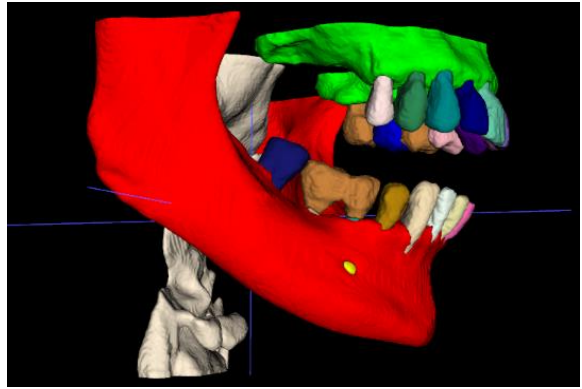

**Fig. 4.** Predictions from the five Phase 1 cross-validation models are ensembled using Multi-Label STAPLE to produce the Phase 1 final ensembled segmentation as seen in this 3D multi-label representation.

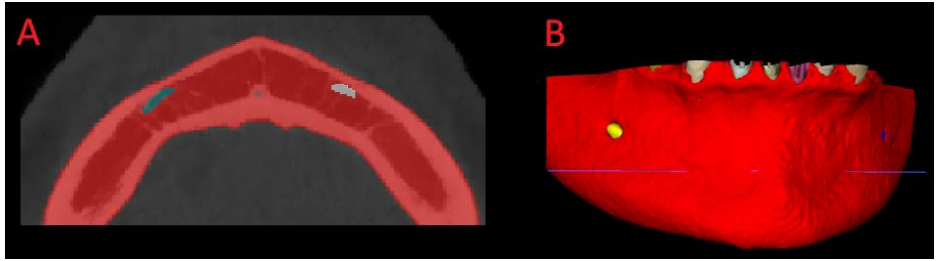

**Fig. 5.** Axial CBCT image (A) and a 3D (B) multi-label representation demonstrating cropping of the Phase 1 ensembled segmentation (Figure 3) performed by identifying the point (x, y, z) representing the most anterior voxel for the easily segmented mandible, then expanding laterally (x) by -110 and +110 voxels; then expanding posteriorly (y) by +100 voxels; then expanding superiorly from the most inferior mandibular point by 90 voxels. This was used to predict the Left Incisive Nerve, Right Incisive Nerve, and Lingual Nerve during Phase 2 inference.

## 3 Results

We evaluated our approach through 5-fold cross-validation on the training set and on the held-out validation set of the ToothFairy3 Challenge (12). **Table 1** summarizes the 5-fold cross-validation performance for the challenge evaluated structures for each fold.

**Table 1.** Cross-validation results and the training set's average volume for each substructure evaluated within the ToothFairy3 challenge dataset.

| Class ID | Structure | Number of Cases Structure Present | Average Volume (mm³) | Fold 1 | Fold 2 | Fold 3 | Fold 4 | Fold 5 | Mean |
|---|---|---|---|---|---|---|---|---|---|
| 0 | Background | — | — | — | — | — | — | — | — |
| 1 | Lower Jawbone | 63 | 46822 | 0.977 | 0.979 | 0.981 | 0.968 | 0.98 | 0.977 |
| 2 | Upper Jawbone | 61 | 12145 | 0.941 | 0.924 | 0.944 | 0.946 | 0.932 | 0.941 |
| 3 | Left Inferior Alveolar Canal | 63 | 410 | 0.793 | 0.809 | 0.825 | 0.793 | 0.837 | 0.793 |
| 4 | Right Inferior Alveolar Canal | 63 | 422 | 0.746 | 0.797 | 0.833 | 0.704 | 0.838 | 0.746 |
| 5 | Left Maxillary Sinus | 44 | 1494 | 0.937 | 0.956 | 0.896 | 0.95 | 0.921 | 0.937 |
| 6 | Right Maxillary Sinus | 42 | 1524 | 0.95 | 0.972 | 0.933 | 0.958 | 0.92 | 0.95 |
| 7 | Pharynx | 63 | 22689 | 0.975 | 0.974 | 0.974 | 0.9 | 0.975 | 0.975 |
| 11 | Upper Right Central Incisor | 59 | 486 | 0.96 | 0.829 | 0.959 | 0.955 | 0.964 | 0.96 |
| 12 | Upper Right Lateral Incisor | 57 | 347 | 0.952 | 0.832 | 0.959 | 0.953 | 0.955 | 0.952 |
| 13 | Upper Right Canine | 58 | 534 | 0.957 | 0.866 | 0.967 | 0.959 | 0.964 | 0.957 |
| 14 | Upper Right First Premolar | 55 | 466 | 0.95 | 0.819 | 0.779 | 0.945 | 0.937 | 0.95 |
| 15 | Upper Right Second Premolar | 50 | 452 | 0.839 | 0.802 | 0.89 | 0.933 | 0.905 | 0.839 |
| 16 | Upper Right First Molar | 51 | 901 | 0.854 | 0.963 | 0.913 | 0.83 | 0.828 | 0.854 |
| 17 | Upper Right Second Molar | 50 | 799 | 0.793 | 0.964 | 0.907 | 0.929 | 0.803 | 0.793 |
| 18 | Upper Right Third Molar (Wisdom Tooth) | 26 | 637 | 0.844 | 0.956 | 0.953 | 0.754 | 0.935 | 0.844 |
| 21 | Upper Left Central Incisor | 57 | 481 | 0.961 | 0.881 | 0.966 | 0.951 | 0.964 | 0.961 |
| 22 | Upper Left Lateral Incisor | 59 | 340 | 0.948 | 0.803 | 0.961 | 0.935 | 0.956 | 0.948 |
| 23 | Upper Left Canine | 59 | 539 | 0.937 | 0.872 | 0.968 | 0.936 | 0.964 | 0.937 |
| 24 | Upper Left First Premolar | 57 | 466 | 0.954 | 0.8 | 0.837 | 0.913 | 0.89 | 0.954 |
| 25 | Upper Left Second Premolar | 53 | 444 | 0.908 | 0.816 | 0.842 | 0.957 | 0.871 | 0.908 |
| 26 | Upper Left First Molar | 54 | 905 | 0.827 | 0.952 | 0.907 | 0.945 | 0.872 | 0.827 |
| 27 | Upper Left Second Molar | 53 | 799 | 0.868 | 0.915 | 0.913 | 0.893 | 0.901 | 0.868 |
| 28 | Upper Left Third Molar (Wisdom Tooth) | 29 | 608 | 0.886 | 0.964 | 0.959 | 0.718 | 0.953 | 0.886 |
| 31 | Lower Left Central Incisor | 59 | 236 | 0.928 | 0.944 | 0.961 | 0.943 | 0.944 | 0.928 |
| 32 | Lower Left Lateral Incisor | 61 | 282 | 0.937 | 0.952 | 0.964 | 0.872 | 0.952 | 0.937 |
| 33 | Lower Left Canine | 60 | 475 | 0.948 | 0.972 | 0.959 | 0.879 | 0.965 | 0.948 |
| 34 | Lower Left First Premolar | 59 | 389 | 0.842 | 0.965 | 0.915 | 0.958 | 0.964 | 0.842 |
| 35 | Lower Left Second Premolar | 55 | 425 | 0.921 | 0.959 | 0.932 | 0.942 | 0.857 | 0.921 |
| 36 | Lower Left First Molar | 40 | 915 | 0.822 | 0.843 | 0.834 | 0.962 | 0.896 | 0.822 |
| 37 | Lower Left Second Molar | 50 | 869 | 0.82 | 0.748 | 0.95 | 0.959 | 0.957 | 0.82 |
| 38 | Lower Left Third Molar (Wisdom Tooth) | 37 | 780 | 0.943 | 0.951 | 0.927 | 0.956 | 0.942 | 0.943 |
| 41 | Lower Right Central Incisor | 60 | 237 | 0.941 | 0.957 | 0.962 | 0.945 | 0.944 | 0.941 |
| 42 | Lower Right Lateral Incisor | 60 | 281 | 0.941 | 0.937 | 0.958 | 0.867 | 0.952 | 0.941 |
| 43 | Lower Right Canine | 59 | 474 | 0.959 | 0.937 | 0.957 | 0.845 | 0.968 | 0.959 |
| 44 | Lower Right First Premolar | 59 | 381 | 0.945 | 0.857 | 0.818 | 0.962 | 0.962 | 0.945 |
| 45 | Lower Right Second Premolar | 55 | 436 | 0.893 | 0.735 | 0.871 | 0.933 | 0.958 | 0.893 |
| 46 | Lower Right First Molar | 45 | 809 | 0.78 | 0.735 | 0.846 | 0.883 | 0.882 | 0.78 |
| 47 | Lower Right Second Molar | 44 | 857 | 0.728 | 0.857 | 0.948 | 0.765 | 0.916 | 0.728 |
| 48 | Lower Right Third Molar (Wisdom Tooth) | 36 | 809 | 0.873 | 0.783 | 0.967 | 0.833 | 0.87 | 0.873 |
| 50 | Tooth Pulp | 61 | 588 | 0.785 | 0.79 | 0.812 | 0.775 | 0.772 | 0.785 |
| 51 | Left Incisive Nerve | 58 | 19 | 0 | 0 | 0.524 | 0.559 | 0.507 | 0.318 |
| 52 | Right Incisive Nerve | 55 | 17 | 0 | 0 | 0 | 0.528 | 0 | 0.106 |
| 53 | Lingual Nerve | 60 | 8 | 0 | 0 | 0 | 0 | 0 | 0 |
| Phase 1 Mean | All Phase 1 Substructures | 63 | — | 0.832 | 0.822 | 0.864 | 0.863 | 0.867 | 0.850 |
| *51 | Left Incisive Nerve Phase 2 | 58 | 19 | 0.689 | 0.663 | 0.702 | 0.697 | 0.691 | 0.688 |
| *52 | Right Incisive Nerve Phase 2 | 55 | 17 | 0.660 | 0.651 | 0.648 | 0.669 | 0.673 | 0.665 |
| *53 | Lingual Nerve Phase 2 | 60 | 8 | 0.680 | 0.691 | 0.677 | 0.674 | 0.682 | 0.681 |
| Combined Mean | Phase 1 Substructures Except Phase 2 Nerves | 63 | — | N/A | N/A | N/A | N/A | N/A | 0.879 |

Each fold's validation Dice was computed as the average Dice across all predicted structures in that fold's validation cases (averaging per-case Dice, weighted equally per class). The scores per fold were relatively consistent, in the range 0.822-0.867. The mean Dice over all folds was approximately 0.850 for these single-model approaches.

We observed that larger structures (teeth and jaw) were segmented with higher accuracy (Dice ~0.977 for mandible), whereas smaller structures like incisive nerves and lingual nerve were more challenging (Dice 0.0-0.318) during Phase 1, lowering the overall Phase 1 average. **Table 1** shows the relationship between 5-fold cross-validation average Dice performance for Phase 1 and Phase 2 substructures vs the average volume of the substructures. Note that the Phase 2 inference successfully improved the performance of the Lingual Nerve Dice from 0.0 to 0.681, the Left Incisive Nerve Dice from 0.318 to 0.688 and the Right Incisive Nerve Dice from 0.106 to 0.665. **Figure 5** shows the 5-fold cross-validation average Dice trend during the Phase 1 training process.

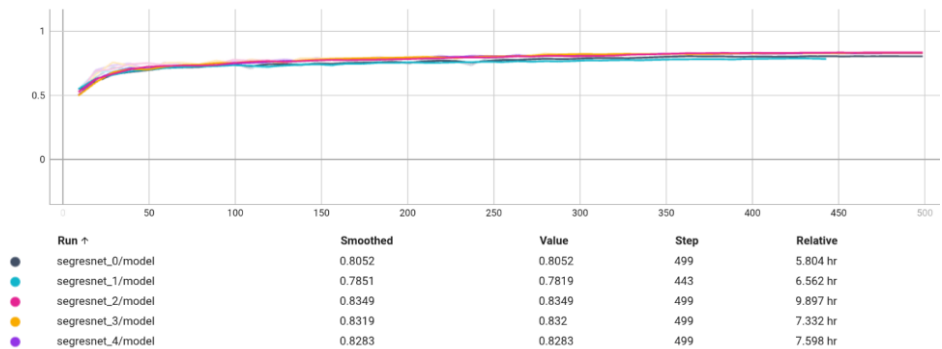

**Fig. 6.** Line plot demonstrating the Phase 1 total training time and validation average dice trends during 5-fold cross validation training since the first validation evaluation conducted at epoch 9 of 500 planned epochs. Note that the second fold (segresnet_1) crashed during epoch 444 of 500.

For the final evaluation on the challenge's out-of-sample validation set, our two-phase ensemble approach achieved an overall average Dice of 0.87. This aggregate performance is slightly higher than the cross-validation mean, which is expected due to the usage of the Multi Label STAPLE ensemble technique, post-processing to clean up the mandible and pharynx labels, and Phase 2's focus on improving the smaller nerve substructures. According to the ToothFairy3 challenge results as of 8/14/25, our method ranked in the upper half of submissions during the debugging/validation phase, indicating potential competitive performance during the final testing phase (12). Qualitatively, the automated segmentations aligned well with the reference standard labels for most structures for in-sample validation cases, as illustrated in **Figure 3**.

The model accurately delineated individual teeth including molar and premolar crowns and roots, even in presence of moderate metal artifacts from dental fillings on in-sample validation qualitative and quantitative analysis. Minor qualitative discrepancies were observed in areas of poor image quality, around metal implants or crowns causing streak artifacts, and the model occasionally missed small portions of a tooth or misclassified an artifact as part of a tooth. Note that the substructure labels referring to the Bridge, Crown, and Implants were not evaluated in the ToothFairy3 challenge ranking metrics, and thus are not reported in this study.

## 4      Discussion

Our challenge submission demonstrates that a light-weight two phase Auto3DSeg framework can produce competitive results for complex multi-class segmentation of dental CBCT scans. By leveraging MONAI's Auto3DSeg framework, we minimized the need for manual network design and hyperparameter tuning. This is particularly useful given the large number of classes (dozens of distinct structures) and the class imbalance inherent in the ToothFairy3 dataset. Larger structures (teeth, jaws) dominate the volume, whereas tiny structures (like canals and nerves) occupy far fewer voxels as shown in **Table 1**. The substructure results demonstrate performance correlating with structure volume as shown in **Table 1**.

The lower Dice for very small classes (nerves) from single phase models on lower resolution [0.6, 0.6, 0.6] mm$^3$ data suggests room for improvement. One possible extension would be additional-phases cropping around specific dental substructures (nerves or pulp) relative to their position to confidently segmented structures (mandible or teeth), similar to our method for focusing on the nerves. This could refine the segmentation of additional fine structures (tooth specific pulp) that a single-phase model might overlook. Notably, this study was successful in improving Lingual Nerve Dice from 0.0 to 0.681, the Left Incisive Nerve Dice from 0.318 to 0.688 and the Right Incisive Nerve Dice from 0.106 to 0.665 after utilizing this multi-phasic cropping-inference with higher resolution of [0.3, 0.3, 0.3] mm$^3$ approach.  Due to time limitations, additional substructure focused phases beyond our nerve-focused Phase 2 phase were not able to be conducted in the present study.

Another point of discussion is the benefit of ensemble fusion using STAPLE (19). The STAPLE algorithm is advantageous in that it accounts for each model's reliability and is theoretically more robust than a simple majority vote or average in cases of systematic bias (19). In our unreported qualitative analysis of the out-of-sample "Set A" and "Set C" test case's inference using STAPLE vs single fold model predictions, STAPLE tended to produce cleaner segmentation borders, especially in areas of uncertainty (for instance, if one fold's model slightly over-segmented a tooth and another fold's model under-segmented it, STAPLE often found a middle ground). This resulted in fewer false positives like isolated tooth fragments. However, STAPLE also introduced a bit of smoothing; occasionally upon qualitative analysis, a tiny structure that only one model detected (e.g. a small root tip fragment) was dropped in the fused result if the other models missed it. In future work, a possible improvement could be to incorporate test-time augmentation or to weight models differently for different subsets of structures (if one model is known to handle nerves better, for example).

An additional strategy to improve on the limited "Set B" dataset consisting of only 63 cases would be to double the size of this dataset by flipping the images and labels laterally along the y/z plane, then swapping all of the label values for the left/right substructures.  This differs from standard random flipping data augmentation, since this also modifies the reference standard labels themselves to their contralateral substructure label values. This was not performed in this study due to time and training time constraints.

From a clinical perspective, the relevance of accurate tooth segmentation in head and neck radiotherapy is significant. With automated segmentation, we can generate dose-volume metrics for each tooth in a patient's radiation plan, something that is impractical to do manually for dozens of teeth. These per-tooth dose metrics could facilitate communication between the oncology and dental team to best inform personalized dental management. For instance, if the model identifies that a particular molar is in a 70 Gy region, the care team might opt for an extraction or intensive prophylactic dental care prior to RT (7). Alternatively, post-radiation, an oral surgeon may reconsider placing an implant into a heavily irradiated region of bone. Conversely, teeth receiving lower doses could be preserved and monitored, avoiding unnecessary extractions. In the long term, the data generated by automated segmentation across many patients could feed into predictive models for ORN or radiation-related dental caries. Previous studies have shown that pre-RT dental care, guided by imaging and dose considerations, can reduce complication rates (5). Automation will make it easier to apply such guidelines consistently. Additionally, our segmentation includes not just the teeth, but also critical adjacent structures (mandible, maxilla, nerves, pulp). This could aid in detection of any anatomic variations, such as an aberrant mandibular canal course, or more refined dose-volume metrics that surgeons and radiation oncologists, respectively, should be aware of when planning interventions (1,3).

One challenge worth noting is the image quality variability in CBCT. Dental CBCT scans often suffer from cone-beam artifacts and scatter, especially in the presence of metal. Our model was trained on the provided dataset, which included typical artifacts, and seemed to generalize across them to an extent. But in cases with extremely poor image quality (e.g., motion blurring or extensive metal streaks), performance may degrade. A potential mitigation strategy is to incorporate metal artifact reduction algorithms or to train the model on simulated artifact-augmented data. Another limitation is that our model did not explicitly differentiate between permanent teeth and dental implants or prosthetic teeth, if present, implants were segmented as generic tooth structures. For radiation planning, this is acceptable, but for dental-specific applications one might want to identify implants separately. The challenge dataset did have a class for implants; however, because it was rare, our model sometimes confused an implant with a tooth root. More targeted training or class weighting could address this in future work. However, as noted earlier, these Bridge, Crown, and Implant structures were not included in the ToothFairy3 challenge ranking metrics.

## 5    Conclusion

We have developed a two-phase 3D multi-class automated segmentation algorithm for dental CBCT images as part of the MICCAI ToothFairy3 Challenge 2025. Our method combined the ease-of-use of Auto3DSeg with a robust Multi Label STAPLE ensemble of a 3D SegResNet model, followed by tight cropping around small and hard to segment substructures. With relatively modest hardware (single GPU, RTX 4090, VRAM requirements < 8 GB), we achieved accurate segmentation of 43 anatomical structures with an overall Dice of 0.87 on the challenge validation set. This performance approaches that of human expert contours for many structures and exceeds earlier atlas-based methods in this domain (5,6). The outcome demonstrates that modern deep learning models can handle the complexity of full-mouth dental segmentation in CBCT, provided that careful preprocessing and training strategies are employed.

Moving forward, we plan to refine the model to further boost accuracy on clinically important substructures, and to integrate the segmentation output into a pipeline for radiation dose analysis in patients with larger field of view head and neck cancer CT simulation scans. The ultimate goal is to deploy such technology in the clinical workflow for head and neck oncology, for example, generating automatic dental reports that flag high-dose teeth and quantify patient-specific risk factors for ORN. With continued improvements, automated tooth segmentation can become a valuable tool to personalize and improve supportive care for patients receiving radiotherapy.

**Data Availability.** Our code for all pre-processing, training, and post-processing are available as open-source at GitHub: dlabella29/ToothFairy25, to facilitate reproducibility and further research by the community. We hope that this work contributes to bridging the gap between computer-assisted dental imaging and practical clinical decision-making, enhancing outcomes in dental and radiation oncology.

**Acknowledgments.** This study was made possible by ToothFairy3 and its organizers. We would like to extend our sincere thanks to the organizers for hosting the challenge and for their continuous support.

**Disclosures of Interests.** The authors have no competing interests to declare that are relevant to the content of this article.